%% file: main.tex
\documentclass[letterpaper, 10 pt, conference]{ieeeconf}  

\IEEEoverridecommandlockouts                              

\title{\LARGE \bf
Shooting for Contact: Contact-Implicit Multiple Shooting for \\Dynamic Motion Retargeting
}
\author{Sergio A. Esteban$^{1}$, Jason H.K. Siu$^{1}$, Derrick Mach$^{1}$, Junheng Li$^{1}$, 
\\
Vince Kurtz$^{2}$, Joel W. Burdick$^{1}$, and Aaron D. Ames$^{1}$%
\thanks{$^{1}$The authors are with the Department of Mechanical and Civil
Engineering, California Institute of Technology, Pasadena, CA, USA.
{\ttfamily\small
\{sesteban,\allowbreak jasonsiu,\allowbreak dmach,\allowbreak
junhengl,\allowbreak ames\}@caltech.edu}
and {\ttfamily\small jwb@robotics.caltech.edu}.}%
\thanks{$^{2}$The author is with the School of Computing, DePaul University,
Chicago, IL, USA. {\ttfamily\small vkurtz1@depaul.edu}.}%
\thanks{*This work was supported by Technology Innovation Institute (TII).}%
}

\input{preamble}

\begin{document}

\maketitle
\thispagestyle{empty}
\pagestyle{empty}

\input{Sections/Abstract}

\input{Sections/Introduction}
\input{Sections/Background}
\input{Sections/Methods}

\input{Sections/Results}

\input{Sections/Conclusion}

\balance

\bibliographystyle{ieeetr}
\bibliography{References/references.bib}

\end{document}

%% file: preamble.tex
\usepackage{amsmath}
\usepackage{amssymb}
\usepackage{amsthm}
\usepackage{mathtools}
\AtBeginDocument{\let\coloneq\coloneqq} 
\usepackage{derivative}
\usepackage{xcolor}
\usepackage{bm}
\usepackage{graphicx}
\usepackage{svg}
\usepackage{subcaption}
\usepackage{mwe}

\usepackage[noadjust]{cite}
\usepackage{url}

\usepackage{graphicx}
\usepackage[export]{adjustbox} 
\graphicspath{{figures/}}

\usepackage{balance}

\usepackage{hyperref} 
\hypersetup{
    colorlinks=true,
    linkcolor=black,
    urlcolor=cyan,
}
\usepackage{booktabs}
\usepackage{pifont} 
\newcommand{\R}{\mathbb{R}}

\newcommand{\mc}[1]{\mathcal{#1}}

\newcommand{\map}[3]{#1\,:\,#2\rightarrow #3}

\newcommand{\bzero}{\mathbf{0}}

\newcommand{\ba}{\mathbf{a}}

\newcommand{\bc}{\mathbf{c}}

\newcommand{\be}{\mathbf{e}}
\newcommand{\bg}{\mathbf{g}}
\newcommand{\bh}{\mathbf{h}}

\newcommand{\bo}{\mathbf{o}}
\newcommand{\bp}{\mathbf{p}}
\newcommand{\bq}{\mathbf{q}}

\newcommand{\bs}{\mathbf{s}}

\newcommand{\bu}{\mathbf{u}}
\newcommand{\bv}{\mathbf{v}}
\newcommand{\bw}{\mathbf{w}}
\newcommand{\bx}{\mathbf{x}}
\newcommand{\by}{\mathbf{y}}
\newcommand{\bz}{\mathbf{z}}

\newcommand{\bB}{\mathbf{B}}

\newcommand{\bF}{\mathbf{F}}

\newcommand{\bH}{\mathbf{H}}

\newcommand{\bJ}{\mathbf{J}}
\newcommand{\bK}{\mathbf{K}}

\newcommand{\bM}{\mathbf{M}}

\newcommand{\bQ}{\mathbf{Q}}
\newcommand{\bR}{\mathbf{R}}

\newcommand{\bU}{\mathbf{U}}

\newcommand{\bX}{\mathbf{X}}

\newcommand{\btheta}{\bm{\theta}}

\newcommand{\btau}{\bm{\tau}}
\newcommand{\bxi}{\bm{\xi}}

\newcommand{\bomega}{\bm{\omega}}
\newcommand{\blambda}{\bm{\lambda}}

\usepackage{blindtext}

\newcommand{\se}[1]{{\color{black} #1}} 

\newcommand{\rev}[1]{{\color{black}#1}} 

%% file: Sections/Abstract.tex
\begin{abstract}
Motion retargeting approaches often prioritize kinematic similarity over whole-body dynamics, contact consistency, and actuation limits, yielding references that are difficult for reinforcement learning (RL) policies to reproduce, particularly for contact-rich behaviors.
We present a contact-implicit, direct simulation-based multiple shooting (DSMS) framework that transforms kinematically feasible references into dynamically feasible whole-body trajectories. 
By embedding a differentiable simulator within a nonlinear program, DSMS resolves contact, friction, impacts, self-collision, and joint limits internally while enforcing tracking, actuation, and task constraints \emph{without} prescribing a contact schedule or introducing explicit contact constraints.
Compared with existing retargeting methods, DSMS accelerates motion-imitation RL training and yields policies with high success rates and low tracking error. We further demonstrate zero-shot sim-to-real transfer on the Unitree G1 through command-conditioned \rev{walking, jogging, and} contact-rich crawling\rev{,} and a highly dynamic 180-degree jump-turn. Additional Material: \href{https://shooting-for-contact.github.io/}
{\texttt{shooting-for-contact.github.io}}.
\end{abstract}


%% file: Sections/Introduction.tex
\section{Introduction}
Reference motions derived from human demonstrations, reduced-order models, and animation provide powerful priors for learning and control \cite{gu2026humanoid}. Recent reference-tracking reinforcement learning (RL) and imitation learning (IL) have demonstrated complex locomotion and whole-body behaviors from motion-capture, animation, and video-generated trajectories \cite{liao2025beyondmimic, peng2018deepmimic, allshirevisual}. 
However, the quality of the learned controller depends strongly on the reference motion \cite{yang2025omniretarget, he2025asap}. Most retargeting pipelines prioritize kinematic correspondence, mapping human pose or key-body poses onto the robot while accounting for differences in morphology \cite{voleti2022smpl, ayusawa2017motion}. Although this \se{approach} can produce visually plausible motions, the resulting trajectories are often not dynamically consistent with robot physics, contacts, or actuation limits.
\begin{figure}[t!]
\vspace{7pt}
    \centering
     \href{https://shooting-for-contact.github.io/}
    {
    \includegraphics[width=1\linewidth]{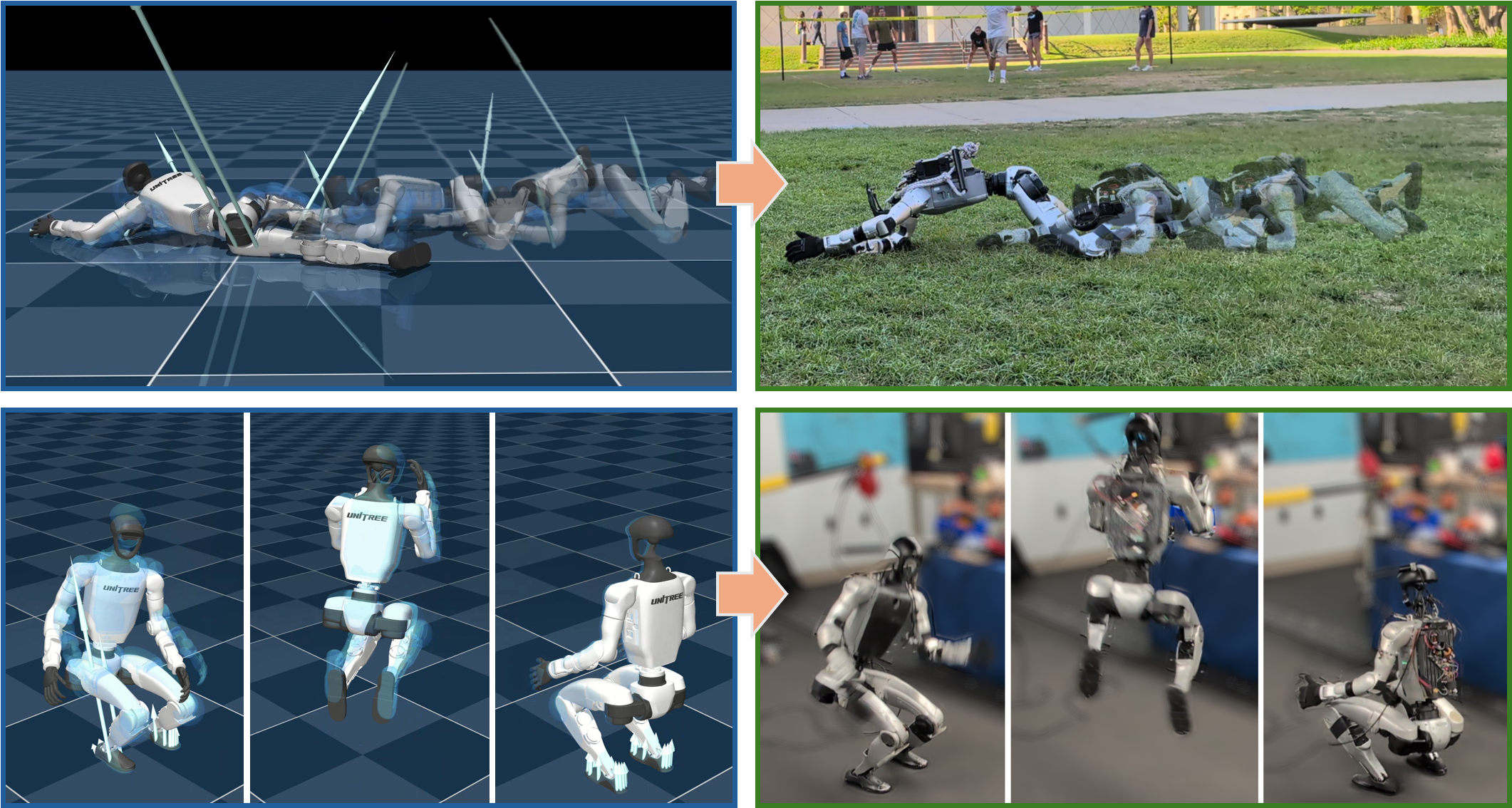}
    }
    \vspace{-15pt}
    \caption{DSMS trajectory optimization solutions in simulation (left) and their corresponding zero-shot hardware realizations on the Unitree G1 (right), shown for velocity-conditioned contact-rich crawling (top) and a $180^\circ$ jump-turn (bottom).}
    \label{fig:hero}
\end{figure}

This dynamics mismatch is particularly problematic for contact-rich behaviors, such as crawling, parkour jumping, and rolling, during which the robot may use different body parts for support and balance \cite{esteban2025reduced}.
A kinematically retargeted motion may contain foot sliding, unrealistic ground penetrations, self-collisions, or actions that require infeasible torques and impulses. RL can sometimes compensate for and robustify these defects, but it must then simultaneously discover a feasible realization of the motion and learn to stabilize it. These characteristics make reward tuning difficult: strong tracking rewards pull the policy toward an infeasible reference, while weaker rewards allow the learned behavior to deviate from the intended motion.

We propose a method for generating \emph{dynamically feasible} references for RL. \rev{Before policy training, trajectory optimization transforms the original motion into a dynamically consistent reference for the target robot, reasoning over contacts not known \textit{a priori} and accommodating arbitrary task constraints. The policy then only needs to stabilize a physically meaningful nominal trajectory rather than discover one from a dynamically inconsistent reference.}
%
%
\subsection{Related Works}
Generating meaningful robot references is challenging because contact-rich humanoid motion is high-dimensional and hybrid. Recent reference-based RL has enabled humanoid robots to reproduce diverse locomotion and whole-body behaviors from human motion data \cite{peng2018deepmimic, liao2025beyondmimic, ze2025twist2}. However, such motions are largely generated through kinematic retargeting and geometric correspondence \cite{wang2026spark}. Other methods improve this stage by allowing interaction-preserving constraints or learned motion refinement \cite{araujo2025retargeting, yang2025omniretarget, zhao2026make}. \rev{Because these pipelines remain largely kinematic, the resulting references can violate the robot's dynamics, degrading the robustness and convergence of downstream tracking policies}~\cite{araujo2025retargeting,liu2025opt2skill}. 

Several works explicitly improve the physical feasibility of reference motions\rev{:} \rev{bi-level} motion imitation alternates between adapting the reference and \rev{the} tracking policy \cite{zhao2025bi}\rev{,} OmniTrack uses privileged policy rollouts \cite{li2026omnitrack}\rev{, and} DynaRetarget \cite{dhedin2026dynaretarget} and DDR \cite{roux2026direct} employ shooting-based trajectory optimization or sampling-based model-predictive control (MPC) \rev{in simulation. However, these approaches either couple reference refinement to policy training or rely on sampling, which complicates enforcing hard task constraints.} \rev{Kinodynamic retargeting~\cite{zhang2026kinodynamic} avoids both but requires a prescribed contact sequence.}

Nonlinear programs based on direct collocation \cite{hereid_HZD} and multiple shooting \cite{olkin2026chasing} have shown promise for stabilizing highly dynamic and underactuated systems. However, in their standard forms, these transcriptions do not resolve contact implicitly and instead require prescribed contact modes or explicit contact constraints.

Contact-implicit trajectory optimization (CITO) provides an appealing basis for ensuring dynamic feasibility \cite{posa2014direct, manchester2019variational}\rev{, since the optimizer is free to make or break contacts as needed while staying close to a reference motion}. However, many CITO methods require explicit contact pairs, complementarity \rev{constraints or} relaxations, \rev{smooth contact models,} or contact-force variables \cite{le2024fast, gu2026humanoid, kim2022contact, sleiman2019contact, kurtz2022contact}. Additionally, many aim primarily at MPC, where full convergence and tight constraints are compromised for speed \cite{le2024fast, kurtz2026inverse}. Related shooting-based methods require differentiable simulations for whole-body MPC, while derivative-free methods such as predictive sampling provide a simple alternative for contact-rich behavior synthesis \cite{zhang2503whole, howell2022predictive}. 
More broadly, differentiable simulators obtain useful sensitivities through regularized contact, implicit differentiation of hard-contact solves, or analytical differentiation of collision and friction \cite{todorov2012mujoco,howell2022dojo,le2023single,lehighly}. 
%
\subsection{Contributions}
This work introduces \textbf{direct simulation-based multi-shooting (DSMS)}\rev{, a nonlinear program transcription for dynamics-aware retargeting of contact-rich humanoid motions in which the discrete transition map of a differentiable simulator defines the whole-body dynamics between shooting nodes. Contact, friction, self-collision, and actuation limits are thus handled internally by the simulator rather than through a traditional contact-implicit formulation. Our main contributions are:}

%
\begin{itemize}
    \item A general direct simulation-based multi-shooting framework for whole-body dynamics and contact-implicit motion retargeting.
    \item A unified pipeline that transforms kinematically feasible trajectories into whole-body dynamically feasible references and that synthesizes command-conditioned periodic gait libraries from a few short motion clips.
    \item For motion-imitation RL, a numerical study of how references generated with reduced-order, kinodynamic, and our full-order optimization affect downstream policy training and tracking performance.
    \item Hardware demonstrations of \rev{(i)} twist command-conditioned \rev{humanoid walking, jogging, and} contact-rich crawling, and \rev{(ii)} a highly dynamic $180^\circ$ jump-turn motion. 
\end{itemize}

%% file: Sections/Background.tex
\section{Preliminaries}
\subsection{Whole-Body Dynamics}
We consider legged robots that achieve locomotion by intermittently making and breaking contact with their environment. In particular, we are interested in contact-rich behaviors in which the robot may use \emph{any} part of its body to generate motion.

We define the state as
$\bx \coloneq (\bq,\bv) \in \mathsf{T}\mc{Q}$, where $\bq~\in~\mc{Q}~\subseteq~\R^{n_q}$ denotes the generalized configuration and $\bv \in \mathsf{T}_{\bq}\mc{Q} \subseteq \R^{n_v}$ the generalized velocity. The whole-body (WB) continuous-time dynamics can be written as
\begin{equation} 
\label{eq:whole_body_dynamics}
    \bM(\bq)\dot{\bv} + \bH(\bq, \bv) = \bB \boldsymbol{\tau} + \bJ_c^{\top}(\bq)\boldsymbol{\lambda},
\end{equation}
where
$\bM : \mc{Q} \rightarrow \R^{n_v \times n_v}$ is the mass--inertia matrix, $\bH : \mathsf{T}\mc{Q} \rightarrow \R^{n_v}$ collects Coriolis, centrifugal, and gravitational terms, $\bB \in \R^{n_v \times n_u}$ maps actuator torques $\btau \in \R^{n_u}$ into generalized coordinates, and $\blambda \in \R^{n_c}$ denotes the stacked contact-force coordinates acting through the contact Jacobian $\bJ_c : \mc{Q} \rightarrow \R^{n_c \times n_v}$. Here, $n_c$ is the total number of scalar contact-force coordinates associated with all active contacts.
%
\subsection{Differentiable Contact Dynamics}
Rigid-body dynamics simulators compute contact forces $\boldsymbol{\lambda}$ and advance the system dynamics \eqref{eq:whole_body_dynamics} with a variety of methods, including compliant models \cite{todorov2012mujoco}, complementarity problems \cite{anitescu1997formulating}, convex optimization \cite{castro2025irrotational}, and more \cite{carpentier2024compliant}. Most simulators provide a discrete-time forward map
\begin{equation}
\label{eq:mujoco_dynamics}
    \bx_{i+1}
    =
    \mathbf{f}(\bx_i,\bu_i)\rev{,}
\end{equation}
where $\bx_i$ and $\bu_i$ are the state and actuator command at the $i^{th}$ simulation time step. A selected actuator interface maps $\bu_i$ to the joint torque $\btau_i$ applied in \eqref{eq:whole_body_dynamics}, as detailed in Sec.~\ref{sec:multi-shooting}. Depending on the simulator, this map may also capture sticking and sliding contacts, impacts, joint-limit activation, and other constraints within a single discrete transition.

We require that these forward dynamics are differentiable, that is, the partial derivatives
\begin{equation} \label{eq:dynamics_grads}
    \frac{\partial \mathbf{f}}{\partial \bx}, \quad \frac{\partial\mathbf{f}}{\partial \bu}
\end{equation}
are available. In this work, we use MuJoCo \cite{todorov2012mujoco}, which leverages a convex differentiable contact formulation \cite{todorov2014convex} to provide these gradients via finite differences. However, our framework is compatible with any differentiable simulator.

%% file: Sections/Methods.tex
\section{Trajectory Optimization} \label{sec:methods}
Our goal is to synthesize dynamically feasible whole-body trajectories that track a desired reference.
Importantly, contact is not modeled \emph{explicitly} in the optimization problem: no contact schedule or complementarity constraints are prescribed. Instead, contact behavior is resolved \emph{implicitly} through the simulator dynamics \eqref{eq:mujoco_dynamics}. 

The tracking terms are defined relative to a reference trajectory $\bX^{\mathrm{ref}} \coloneq (\bx_0^{\mathrm{ref}},\ldots,\bx_N^{\mathrm{ref}})$ that need not satisfy the whole-body dynamics \eqref{eq:whole_body_dynamics}. The reference may come from retargeted human motion-capture data, animation data, or a reduced-order model (ROM), whose trajectories are dynamically consistent only within the reduced representation.

\subsection{Direct Simulation-based Multiple Shooting (DSMS)} \label{sec:multi-shooting}
We employ direct multiple shooting to transcribe the trajectory optimization problem into a finite-dimensional nonlinear program (NLP).
This transcription divides the horizon into $N$ intervals, treats the shooting-node states as decision variables, and enforces continuity between independently integrated intervals through defect constraints. \rev{Shorter rollouts reduce long-horizon sensitivity and improve convergence~\cite{wensing2023optimization}, which is particularly valuable for highly dynamic, underactuated motions whose open-loop behavior is sensitive to early control decisions.}
%

We discretize the horizon $[0,T]$ into $N$ shooting intervals on the uniform grid
\begin{equation}
\nonumber
    t_k \coloneq k\,\Delta t, \qquad k = 0,\ldots,N, \qquad
    \Delta t \coloneq T/N.
\end{equation}
Let $\bX \coloneq (\bx_0,\ldots,\bx_N)$ collect the state variables at the shooting nodes and let $\bU \coloneq (\bxi_1,\ldots,\bxi_M)$ collect $M$ control points, with $\bxi_j \in \R^{n_u}$. These control points parameterize the time-varying command signal through the map \mbox{$\map{\bs}{[0,T]\times\R^{Mn_u}}{\R^{n_u}}$}, defined by
\begin{equation}
\label{eq:spline_command}
    \bu(t)
    \coloneq
    \bs(t,\bU)
    =
    \sum_{j=1}^{M}
    \beta_j(t)\bxi_j.
\end{equation}
Here, $\{\beta_j\}_{j=1}^{M}$ \rev{are prescribed basis functions, such as zero-order hold or piecewise linear. The actuator command $\bu(t)$ is either a joint torque or a desired joint position tracked by a low-level PD controller,}\footnote{Under torque control, the applied torque is $\btau(t)=\bu(t)$ and under position control, the applied torque is $\btau(t)=\bK_p\bigl(\bu(t)-\bq^\mathrm{jnt}(t)\bigr)-\bK_d\bv^\mathrm{jnt}(t)$.} \rev{and $\bu_k \coloneq \bu(t_k)$ at node $k$.}
\begin{figure}[t!]
    \centering
    \includegraphics[width=0.7 \linewidth]{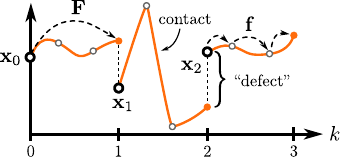}
    \centering
    \caption{Proposed multi-shooting trajectory optimization.}
    \label{fig:multi-shooting}
    \vspace{-2pt}
\end{figure}
%
Using shooting-node states $\bX$ and spline control points $\bU$, we transcribe the optimal control problem via direct multiple shooting as an NLP:
\begin{subequations}
\label{eq:ocp}
\begin{align}
    \min_{\bX,\bU}\quad
    & J(\bX,\bU)
    \label{eq:ocp_objective} \\
    \mathrm{s.t.}\quad
    & \bx_{k+1} = \bF(\bx_k,\bu_k),
      && k=0,\ldots,N-1,
    \label{eq:ocp_dynamics} \\
    & \bg(\bx_k,\bu_k) \leq \bzero,
    \label{eq:ocp_inequality} \\
    & \bh(\bx_k,\bu_k) = \bzero.
    \label{eq:ocp_equality}
\end{align}
\end{subequations}
Here, the objective~\eqref{eq:ocp_objective} encodes tracking and regularization, \rev{and}~\eqref{eq:ocp_inequality} and~\eqref{eq:ocp_equality} \rev{collect inequality and equality constraints, such as path limits, boundary conditions, or task constraints}. The flow map $\bF$ in~\eqref{eq:ocp_dynamics} integrates~\eqref{eq:mujoco_dynamics} from $\bx_k$ under $\bu(t)$ during $[t_k,t_{k+1})$, with the \emph{defect} enforcing continuity between shooting intervals\rev{, as illustrated in Fig.~\ref{fig:multi-shooting}}. Internally, $\bF$ chains $S$ substeps of $\mathbf{f}$ with $\Delta t_{\mathrm{sim}}=\Delta t/S$\rev{, so its derivatives with respect to the decision variables follow from chaining the per-substep derivatives~\eqref{eq:dynamics_grads}}. Fine simulator steps resolve stiff contact dynamics, while state variables only at the coarser shooting nodes keep the NLP small. Because the simulator directly defines the flow between shooting nodes, we refer to this formulation as \emph{direct simulation-based multiple shooting (DSMS)}.
%
\subsection{Cost Function} \label{sec:cost_function}
The cost $J(\bX,\bU)$ in \eqref{eq:ocp_objective} is a weighted sum of state and body tracking, together with input regularizers penalizing torque effort and command rate.

With state error $\be_k \coloneq \bx_k \ominus \bx_k^{\mathrm{ref}}$, the state tracking term\footnote{Subtraction on a manifold is denoted by $\ominus$, accounting for the quaternion component of the state.} penalizes the tangent space deviation from the reference,
\begin{equation}
\label{eq:cost_state}
    \ell_{\bx}(\bx_k, k) = \be_k^\top \bQ_\bx \, \be_k,
\end{equation}
with diagonal $\bQ_\bx  \succeq \bzero$ weighting the configuration and velocity errors. 
%

In the style of the motion tracking objectives used in \cite{liao2025beyondmimic, peng2018deepmimic}, we track the pose and twist of a set of key bodies  $\mc{B}$, such as the feet, hands, and pelvis. For body $b \in \mc{B}$, let $\by^b_k$ stack its pose and twist,
\begin{equation}
\label{eq:body_output}
    \by^b_k
    \coloneq
    \bigl(\,
        \underbrace{\bp^b_k,\ \boldsymbol{q}^b_k}_{\text{pose}},\ \
        \underbrace{\bv^b_k,\ \bomega^b_k}_{\text{twist}}
    \,\bigr), 
\end{equation}
where these quantities are obtained by forward kinematics from $\bx_k$, and let $\by^{b,\mathrm{ref}}_k$ denote the same quantities on $\bx^{\mathrm{ref}}_k$.  With body error $\be^b_k \coloneq \by^b_k \ominus \by^{b,\mathrm{ref}}_k$, the body tracking term is
\begin{equation}
\label{eq:cost_body}
    \ell_{\by}(\bx_k, k) = \sum_{b\in\mc{B}} \bigl(\be^{b}_k \bigr)^{\top} \bQ_\by \be^b_k,
\end{equation}
with diagonal $\bQ_\by \succeq \bzero$ weighting each body's pose and twist errors. Each body is tracked in either the world frame or the frame of an anchor body such as the torso. 
%

%
\begin{figure*}[t!]
    \centering
    \begin{subfigure}[c]{0.35\textwidth}
        \centering
        \includegraphics[width=0.95\linewidth]{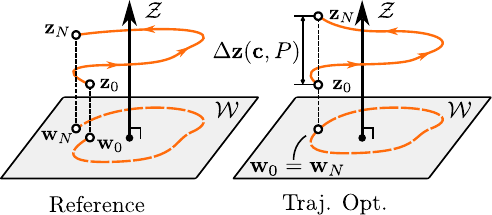}
        \caption{Limit cycle closure.}
    \end{subfigure}%
    \hspace{5pt}%
    \begin{subfigure}[c]{\dimexpr0.65\textwidth-5pt\relax}
        \centering
        \includegraphics[
            width=0.95\linewidth,
            keepaspectratio
        ]{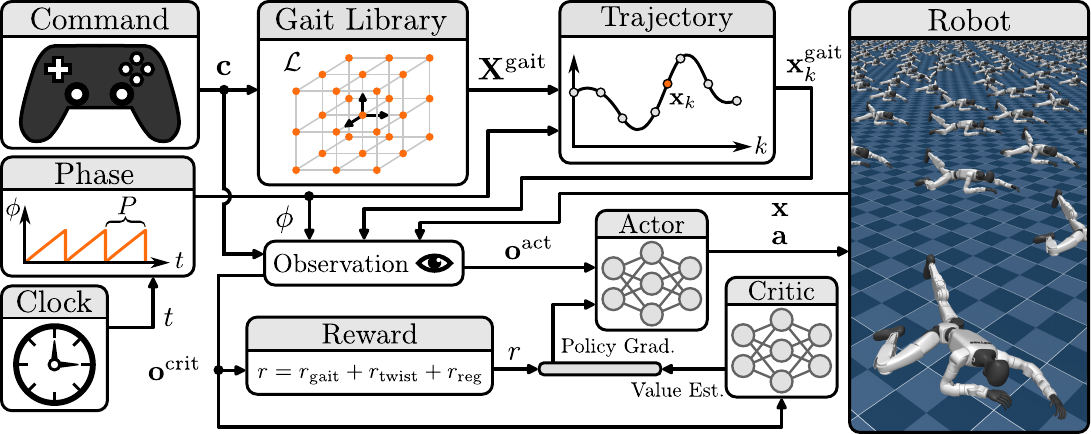}
        \caption{Learning architecture.}
    \end{subfigure}
    \caption{(a) A gait reference is closed into a command-constrained limit cycle: the planar pose advances by $\Delta\bz(\bc\rev{, P})$, while $\bw_0=\bw_N$ enforces limit-cycle closure. (b) The commanded twist selects a phase-aligned gait from the library, and an asymmetric actor–critic trains a reference-free policy from proprioception, command, and phase.}
    \label{fig:gait_and_learning}
    \vspace{-20pt}
\end{figure*}

The terminal state and body  tracking terms retain the same form but with their weights scaled by $\gamma > 1$:
%
\begin{subequations}
    \begin{align}
        \ell_{\bx}^{f}(\bx_N) &= \gamma \, \ell_{\bx}(\bx_N,N),
        \\
        \ell_{\by}^{f}(\bx_N) &= \gamma \,\ell_{\by}(\bx_N,N).
    \end{align}
\end{subequations}

Because the state and body errors are defined in different spaces---a generalized state space and a body-output space---the two terms are complementary: body tracking enforces task-relevant motion, while state tracking resolves kinematic redundancy and regularizes the whole-body posture.

We regularize actuation with two terms. The first penalizes torque effort on the realized torque $\btau_k$,
\begin{equation}
\label{eq:cost_torque}
    \ell_{\btau}(\bx_k, \bu_k) = \btau_k^\top \bR_{\btau} \, \btau_k,
\end{equation}
with diagonal $\bR_{\btau} \succeq \bzero$. We penalize the realized torque rather than the raw command $\bu_k$ because the meaning of $\bu_k$ depends on the actuator interface. The second term penalizes the command difference $\Delta \bu_k = \bu_k - \bu_{k-1}$, promoting smooth commands and suppressing chatter,
\begin{equation}
\label{eq:cost_input_rate}
    \ell_{\Delta \bu}(\bU) = \sum_{k=1}^{N-1} \Delta \bu_k^\top \bR_{\Delta \bu}\, \Delta \bu_k,
\end{equation}
where diagonal $\bR_{\Delta \bu} \succeq \bzero$ and $\bu_k = \bs(t_k, \bU)$.

Collecting the tracking and regularization terms, the objective $J(\bX,\bU)$ in \eqref{eq:ocp_objective} is
\begin{align}
    J(\bX,\bU) =\, &
    \Delta t \sum_{k=0}^{N-1}
    \Bigl[
    \ell_{\bx}(\bx_k,k)
    + \ell_{\by}(\bx_k,k)
    + \ell_{\btau}(\bx_k,\bu_k)
    \Bigr]
    \notag
    \\
    &+
      \ell_{\bx}^{f}(\bx_N)
    + \ell_{\by}^{f}(\bx_N)
    + \Delta t \, \ell_{\Delta\bu}(\bU),
\label{eq:cost_total}
\end{align}
where the running terms are scaled by the timestep $\Delta t$ so that their sum approximates the integral cost.

\subsection{Receding Horizon Control} \label{sec:mpc}

Full-horizon trajectory optimization solves the complete reference trajectory from a fixed initial state in a single solve. For highly dynamic humanoid motions, such as backflips and \rev{side rolls (Fig.~\ref{fig:simulations})}, we instead apply~\eqref{eq:ocp} in a receding-horizon fashion\rev{. Full-horizon solves converge less reliably for such long, highly dynamic motions, whereas short replanning windows remain tractable and still yield a dynamically feasible rollout}. Each solve spans a moving reference window of $H$ intervals and duration $T_H = H\Delta t$, shorter than the full trajectory.

At the $j$-th replanning time $\bar{t}_j$, we initialize the horizon from the current simulated state and shift the reference accordingly:
\begin{equation}
    \bx_0 = \bx(\bar{t}_j), \qquad
    \bx_k^{\mathrm{ref}}
    =
    \bx^{\mathrm{ref}}(\bar{t}_j + k\Delta t),
\end{equation}
for $k=0,\ldots,H$.
We then solve the corresponding $H$-interval horizon instance of~\eqref{eq:ocp}, execute the first \rev{$\Delta t_{\mathrm{rhc}}$} seconds of the optimized command, advance the simulator, and repeat with the shifted window. Concatenating the executed segments produces a long rollout whose transitions are generated directly by the simulator and therefore satisfy~\eqref{eq:whole_body_dynamics} by construction, rather than only up to the defect tolerance in~\eqref{eq:ocp_dynamics}.
%
\subsection{Gait Synthesis} \label{sec:gait_synthesis}
To demonstrate that the proposed optimization can enforce constraints while preserving dynamic feasibility, we \rev{synthesize libraries of periodic walking, jogging, and crawling gaits from human motion-capture data, where each entry realizes} a prescribed average base twist $\bc \coloneq (v^x,v^y,\omega^z)\in\R^3$. \rev{Notably, crawling involves} frequent contacts with the hands, elbows, knees, and feet, including sliding contacts. \rev{For each gait, forward, backward, and turning references, together with a static, zero-twist idle reference (standing or prone), tile the command space.}

\begin{figure*}[t]
    \centering
    \begin{subfigure}[t]{\dimexpr(\linewidth-8pt)/3\relax}
        \centering
        \includegraphics[width=\linewidth]{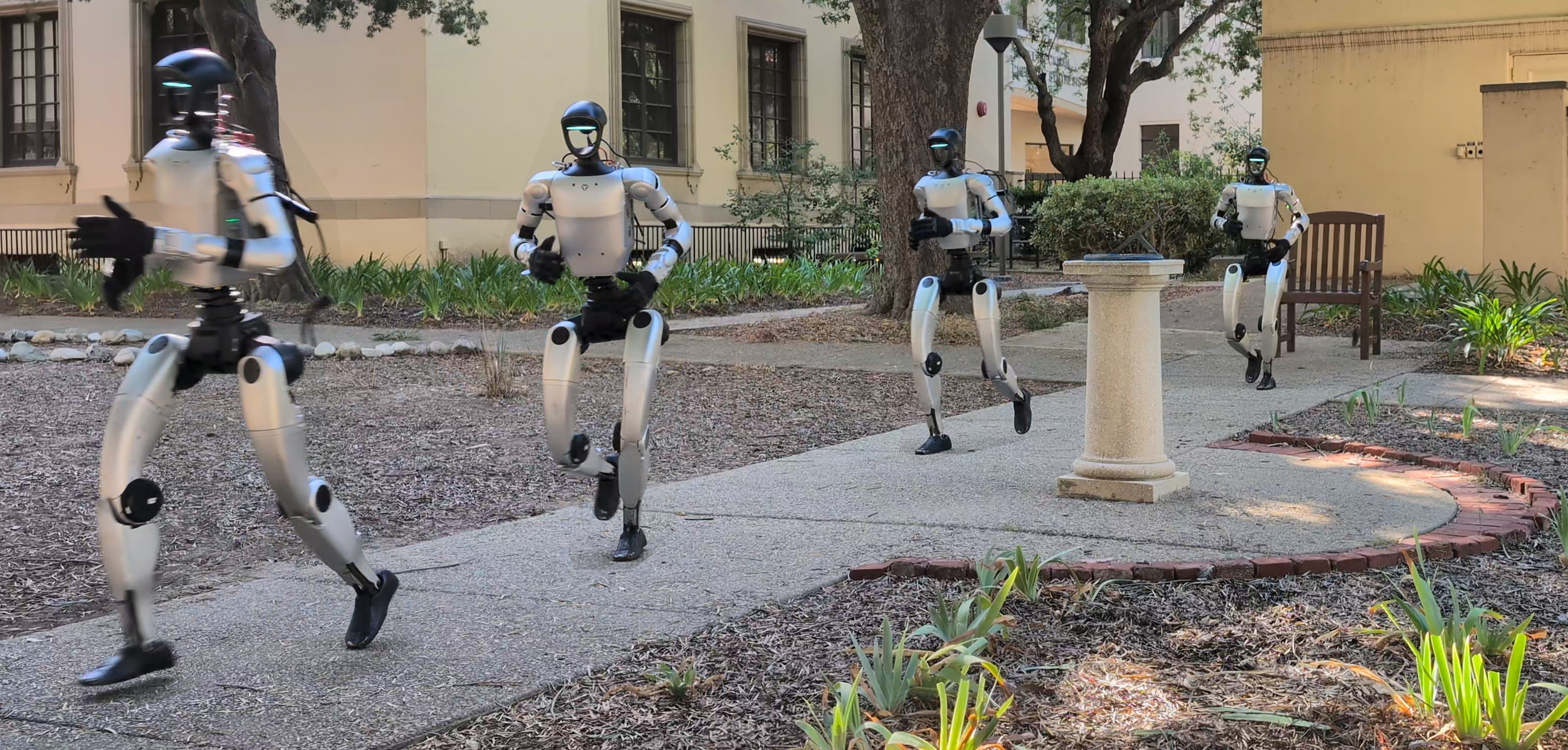}
        \caption{}
        \label{fig:result_a}
    \end{subfigure}\hfill
    \begin{subfigure}[t]{\dimexpr(\linewidth-8pt)/3\relax}
        \centering
        \includegraphics[width=\linewidth]{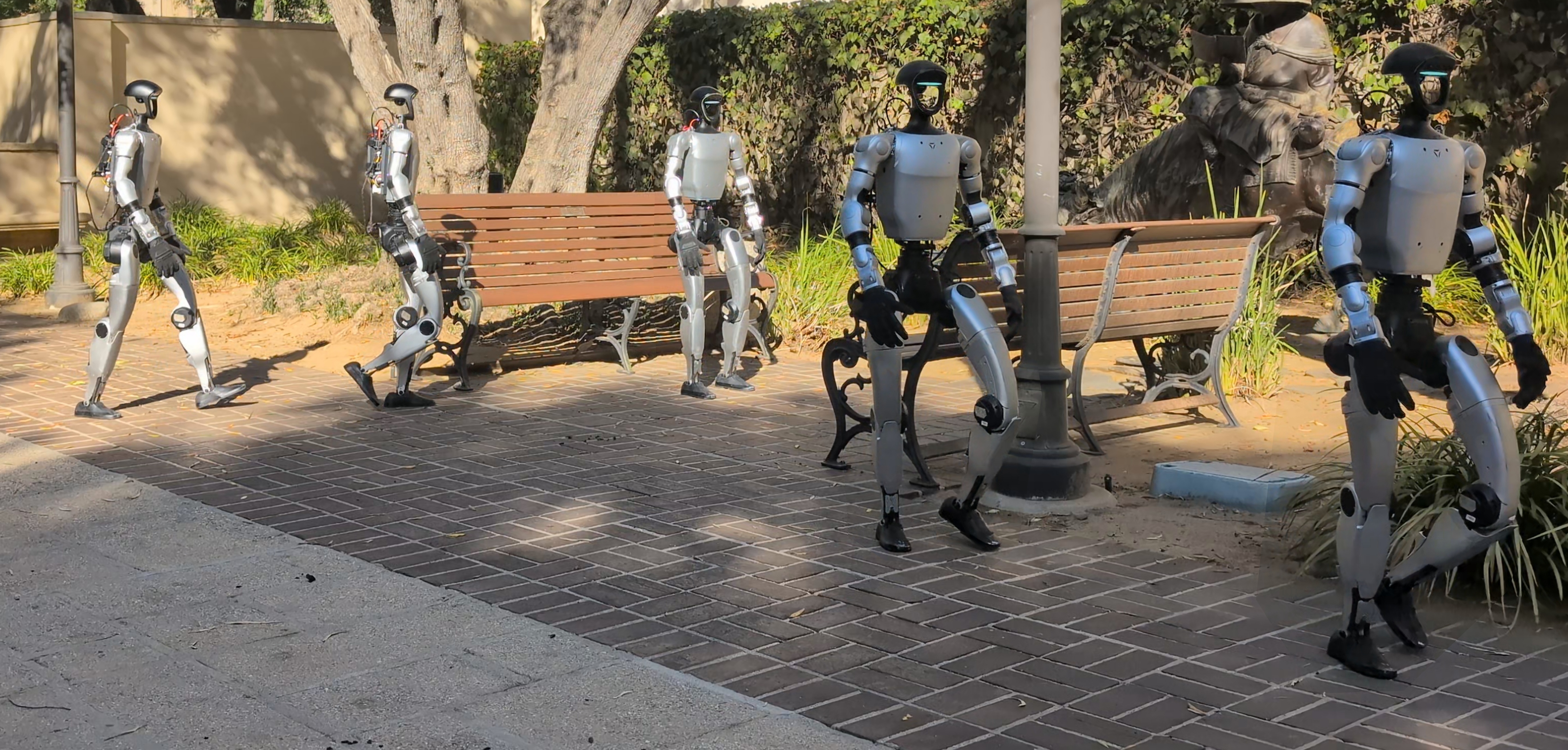}
        \caption{}
        \label{fig:result_b}
    \end{subfigure}\hfill
    \begin{subfigure}[t]{\dimexpr(\linewidth-8pt)/3\relax}
        \centering
        \includegraphics[width=\linewidth]{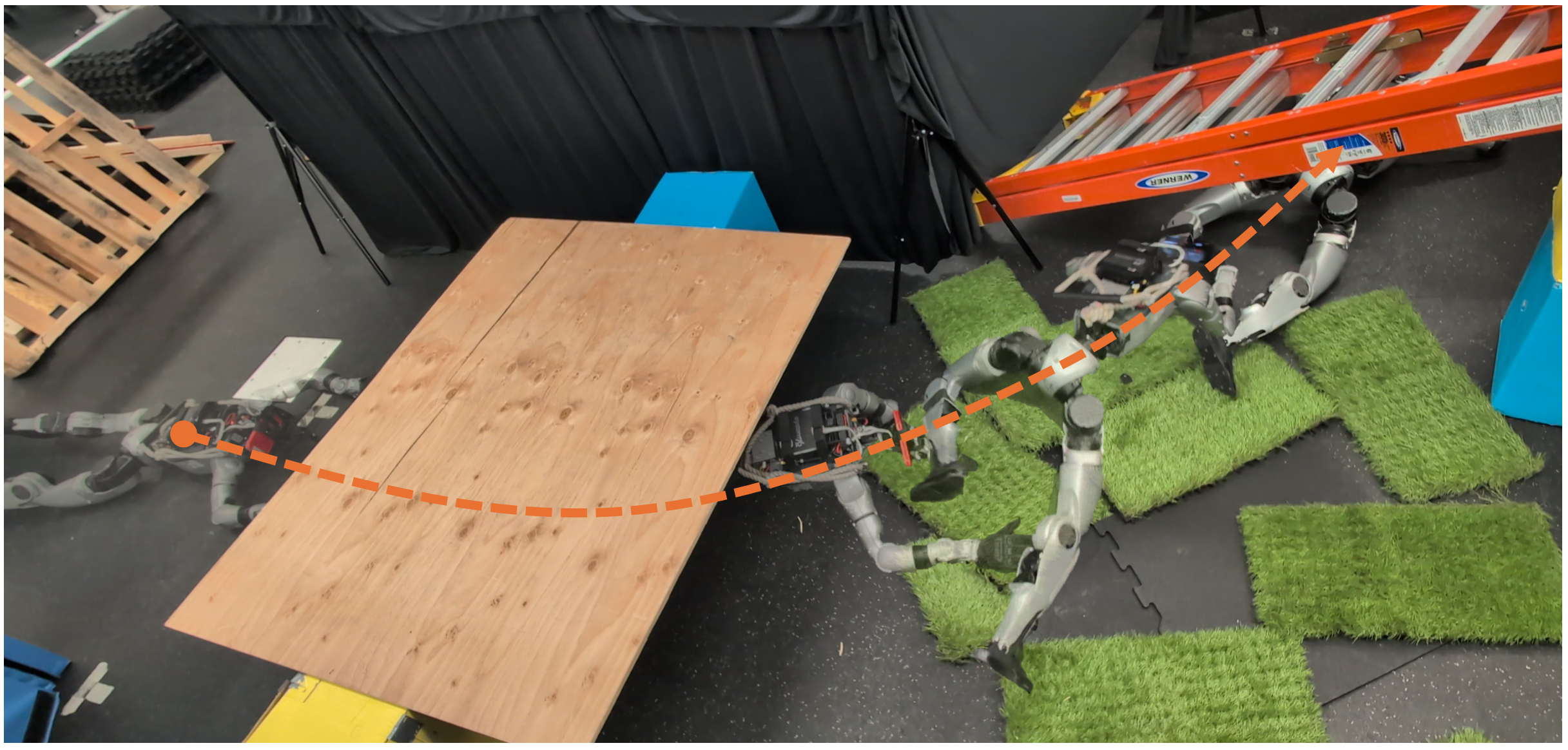}
        \caption{}
        \label{fig:result_c}
    \end{subfigure}
    \vspace{-6pt}
    \caption{Experimental demonstrations of (a) jogging forward, (b) walking while turning, and (c) crawling under height constrained spaces with rough patches.}
    \label{fig:hardware3}
    \vspace{-10pt}
\end{figure*}
Starting from a nearly periodic gait reference with period $P$, we solve the NLP~\eqref{eq:ocp} over a single gait cycle, using the equality constraints~\eqref{eq:ocp_equality} to enforce limit cycle closure. To formulate these constraints, we split the shooting-node state as $\bx_k = (\bz_k,\ \bw_k) \in \mc{Z}\times\mc{W}$. The first component is the planar base pose,
\begin{equation}
\label{eq:planar_pose}
    \bz_k \coloneq \bigl(\bp^{xy}_k,\ \psi_k\bigr) \in \mc{Z},
\end{equation}
containing the base horizontal position $\bp^{xy}_k$ and heading $\psi_k$, with $\mc{Z}\cong SE(2)$. The second component is the remaining state,
\begin{equation}
\label{eq:periodic_state}
    \bw_k \coloneq \bigl(p^z_k,\ \btheta_k,\ \bq^{\mathrm{jnt}}_k,\ \bv_k\bigr) \in \mc{W},
\end{equation}
containing the base height $p^z_k$, yaw-invariant tilt $\btheta_k$, joint positions $\bq^{\mathrm{jnt}}_k$, and generalized velocities $\bv_k$.

The planar pose advances by the commanded displacement,
$\bz_N = \bz_0 \oplus \Delta\bz(\bc, P)$,\footnote{Addition on a manifold is denoted by $\oplus$ and accounts for the $SE(2)$ component of the state.} which enforces a prescribed average base twist $\bc$. The remaining state satisfies $\bw_0 = \bw_N$, returning the height, tilt, joints, and velocities to their initial values and enforcing periodicity. Together, these constraints yield the limit cycle shown in Fig.~\ref{fig:gait_and_learning}a. Because closure determines the cycle only up to a planar rigid transformation, the tracking costs~\eqref{eq:cost_state} and~\eqref{eq:cost_body} anchor it to the gait reference, fully determining the remaining degrees of freedom.

In addition to the cost~\eqref{eq:cost_total}, we add a no-slip term that penalizes the horizontal velocity of each contact link while it is planted. 
Because this term is a soft penalty rather than a hard constraint, it discourages unnecessary slipping while permitting the sliding contacts required by crawling, without excluding dynamically feasible solutions.

Repeating this procedure for a range of commanded twists yields a library of command–gait pairs, in which each gait trajectory $\bX^{\mathrm{gait}}_i$ is dynamically feasible under its corresponding command $\bc_i$:
\begin{equation}
    \mc{L} =
    \{ 
    (\bc_i,\bX^{\mathrm{gait}}_i) 
    \}_{i=1}^{N_{\mathrm{gait}}}.
\end{equation}
%
\section{Reinforcement Learning}
\subsection{Motion Imitation} \label{sec:rl_imitation}
\rev{Following the motion-imitation framework of~\cite{liao2025beyondmimic}, we train RL policies to reproduce the generated reference motions.} Given a reference trajectory from Sec. \ref{sec:mpc}, the policy learns to reproduce it using per-frame body pose and twist rewards, reference-state initialization, and adaptive frame sampling.

\rev{Simulation deployment then measures} how readily and accurately each reference can be learned and reproduced: a policy is trained on each reference and evaluated by how reliably and accurately it reproduces the motion during simulation rollouts. \rev{Sec.~\ref{sec:ablation_comp} uses these results} to compare dynamic fidelity and retargeting methods.

\subsection{Locomotion Controller} \label{sec:rl_locomotion}
We train a single command-conditioned RL policy to track the gait library $\mc{L}$. Given a commanded twist $\bc$, the nearest library trajectory provides the imitation target, allowing the discrete library to serve as a continuous velocity-command interface, as shown in Fig.~\ref{fig:gait_and_learning}b.

We use an asymmetric actor-critic with a reference-free actor that observes signals available at deployment:
%
\begin{equation}
    \bo^{\mathrm{act}} =
    \big(
    \hat{\bg}, \;
    \boldsymbol{\omega}^{\mathrm{base}},\;
    \bq^{\mathrm{jnt}}, \; 
    \bv^{\mathrm{jnt}},\;
    \ba^{\mathrm{prev}}, \bc,\; 
    \sin(\phi), \; \cos(\phi)
    \big).
\end{equation}
These comprise projected gravity $\hat{\bg}$, base angular velocity $\boldsymbol{\omega}^{\mathrm{base}}$, joint positions and velocities $\bq^{\mathrm{jnt}},\bv^{\mathrm{jnt}}$, previous action $\ba^{\mathrm{prev}}$, commanded twist $\bc$, and a sinusoidal encoding of the gait phase $\phi$. During training, the critic additionally observes the base linear velocity\rev{, key-body poses $(\bp^{\mc{B}},\bq^{\mc{B}})$,} and gait reference:
\begin{equation}
    \bo^{\mathrm{crit}} =
    \big(
    \bo^{\mathrm{act}}, \;
    \bv^{\mathrm{base}}, \;
    \rev{\bp^{\mc{B}}, \; \bq^{\mc{B}},} \;
    \bx^{\mathrm{gait}}(\bc, \phi)
    \big).
\end{equation}
Here, $\bx^{\mathrm{gait}}(\bc,\phi)$ is the reference state for command $\bc$ at phase $\phi$. 
The actor therefore requires only proprioception, command, and phase at inference.

The action $\ba$ specifies joint-position offsets applied through a PD controller. We use the \rev{gait's} idle posture \rev{(standing or prone)} as the nominal pose, so zero action is valid and the policy learns only residual corrections.

The reward consists of gait tracking, command tracking, and regularization:
\begin{equation}
r = r_{\mathrm{gait}} + r_{\mathrm{twist}} + r_{\mathrm{reg}}.
\end{equation}
Following \cite{liao2025beyondmimic}, $r_{\mathrm{gait}}$ tracks key-body poses and velocities, together with the anchor position and \rev{orientation}. The anchor target is propagated egocentrically by the reference \rev{anchor velocity} and re-based at each command change to preserve translation and turning across gait cycles. Finally, $r_{\mathrm{twist}}$ tracks the commanded base twist, while $r_{\mathrm{reg}}$ penalizes \rev{undesirable behavior such as jerky actions or unsafe contacts}.

All trajectories in $\mc{L}$ share period $P$, so a phase clock $\phi$ indexes the current imitation target. During an episode, resampling $\bc$ selects the nearest gait without resetting the phase or robot state, while $\bc=\bzero$ selects the idle reference. At each switch, we blend the outgoing and incoming references at the shared phase while updating the command immediately, enabling smooth transitions.

%% file: Sections/Results.tex
\begin{figure*}[!t]
\vspace{0.2cm}
    \center	
    \includegraphics[clip, trim=0.2cm 9.2cm 0.2cm 0.0cm, width=2\columnwidth]{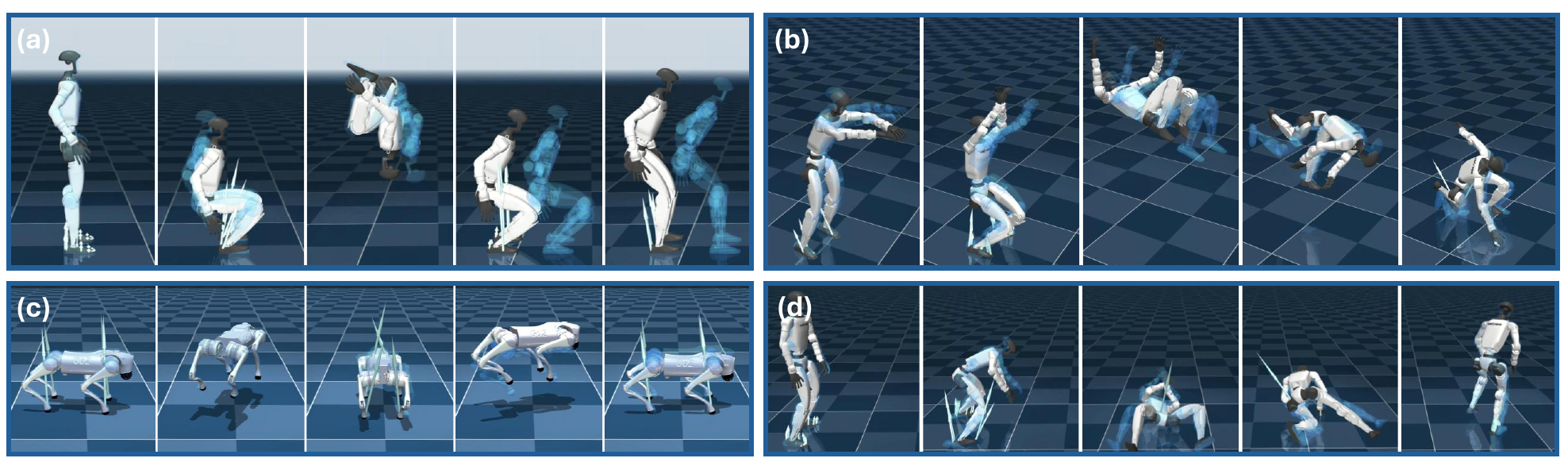}
    \caption{Simulation snapshots for dynamic contact-rich whole-body maneuvers: (a) humanoid backflip, (b) humanoid super hero backflip, (c) quadruped jump-turn, and (d) humanoid side-rolling. Blue ghost visualizes dynamically infeasible references.}
    \label{fig:simulations}
    \vspace{-0.4cm}
\end{figure*}


\section{Results} \label{sec:results}

\subsection{Implementation Details}

\rev{We solve~\eqref{eq:ocp} with the interior-point solver IPOPT~\cite{wachter2006implementation} via \texttt{cyipopt}, supplying the cost, constraints, and their first derivatives. IPOPT forms the NLP Lagrangian internally and factorizes each sparse KKT system with HSL \texttt{ma57}, exploiting the block-sparse multiple-shooting Jacobian. Since MuJoCo does not provide the second-order derivatives of~\eqref{eq:mujoco_dynamics} required by the Lagrangian Hessian,\footnote{Second-order derivatives are available through automatic differentiation in MJX, MuJoCo's JAX-based reimplementation.} we use IPOPT's limited-memory (L-BFGS) Hessian approximation built from the first-order derivatives~\eqref{eq:dynamics_grads}. This approximation performs surprisingly well despite the stiff contact dynamics and abrupt curvature changes at contact events.}

We train RL policies for the Unitree G1 in mjlab~\cite{zakka2026mjlablightweightframeworkgpuaccelerated} with reference clips from the BONES-SEED human motion dataset~\cite{bonesstudio2025bonesseed}.\footnote{\rev{Motion Data by Bones Studio -- \url{https://bones.studio/}.}} Policies are optimized with PPO via \texttt{rsl\_rl}, using $(512,256,128)$ actor and critic MLPs. Simulated dynamics run at $200$~Hz and the policy at $50$~Hz, with $10$~second episodes. For sim-to-real transfer, we randomize base pushes, center-of-mass offsets, joint-encoder biases, and contact friction.
\subsection{Hardware Deployment}
We first deploy a motion imitation policy trained to reproduce a $180^\circ$ jump-turn reference from a single rigid-body (SRB) trajectory. The controller executes this highly dynamic maneuver by rotating the robot by approximately $180^\circ$ during the jump and recovering to a stable landing (Fig.~\ref{fig:hero}). We next deploy the RL \rev{locomotion controllers} (Sec.~\ref{sec:rl_locomotion}) \rev{for walking, jogging, and crawling, steered} in real time through commanded twists. \rev{The robot crawls outdoors on grass (Fig.~\ref{fig:hero}). It also jogs forward and walks while turning outdoors, and crawls through a height-constrained indoor course with rough patches} (Fig.~\ref{fig:hardware3}). \rev{The walking, jogging, and crawling references are generated with full-horizon trajectory optimization (TO) and the jump-turn with receding-horizon control (RHC), using the parameters in Tables~\ref{tab:params_to} and~\ref{tab:params_rhc}. All motions use a piecewise-linear control basis.}

\rev{Notably, every policy transfers \emph{zero-shot} to the Unitree G1 hardware, without any real-world fine-tuning.}

\begin{table}[t]
    \centering
    \caption{\rev{TO parameters for the walking, jogging, and crawling gaits.}}
    \label{tab:params_to}
    \footnotesize
    \setlength{\tabcolsep}{4pt}
    \setlength{\aboverulesep}{0pt}
    \setlength{\belowrulesep}{0pt}
    \renewcommand{\arraystretch}{1.15}
    \begin{tabular}{l|cccccc}
        \toprule
        \rev{\textbf{Motion}} & \rev{$\Delta t_{\mathrm{sim}}$ [ms]} & \rev{$\Delta t$ [ms]} & \rev{$S$} & \rev{$P$ [s]} & \rev{$N$} & \rev{$M$} \\
        \midrule
        \rev{Walk} & \rev{10} & \rev{20} & \rev{2} & \rev{1.40} & \rev{70} & \rev{71} \\
        \rev{Jog} & \rev{5} & \rev{20} & \rev{4} & \rev{0.86} & \rev{43} & \rev{44} \\
        \rev{Crawl} & \rev{5} & \rev{20} & \rev{4} & \rev{2.60} & \rev{130} & \rev{131} \\
        \bottomrule
    \end{tabular}
\end{table}

\begin{table}[t]
    \centering
    \caption{\rev{RHC parameters for the $180^\circ$ jump-turn.}}
    \label{tab:params_rhc}
    \footnotesize
    \setlength{\tabcolsep}{4pt}
    \setlength{\aboverulesep}{0pt}
    \setlength{\belowrulesep}{0pt}
    \renewcommand{\arraystretch}{1.15}
    \begin{tabular}{l|cccccc}
        \toprule
        \rev{\textbf{Motion}} & \rev{$\Delta t_{\mathrm{sim}}$ [ms]} & \rev{$\Delta t$ [ms]} & \rev{$S$} & \rev{$\Delta t_{\mathrm{rhc}}$ [ms]} & \rev{$H$} & \rev{$M$} \\
        \midrule
        \rev{Jump-turn} & \rev{10} & \rev{20} & \rev{2} & \rev{20} & \rev{30} & \rev{30} \\
        \bottomrule
    \end{tabular}
\end{table}
\subsection{Locomotion Velocity Tracking}
\begin{figure}[t]
    \centering
    \includegraphics[width=1.0 \linewidth]{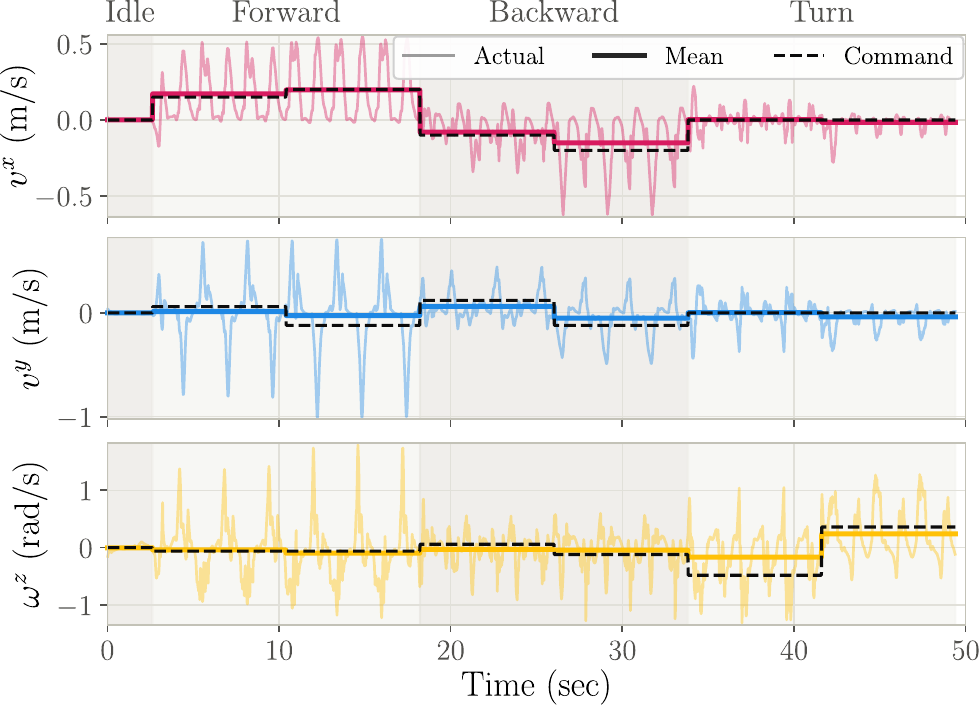}
    \centering
    \vspace{-12pt}
    \caption{Commanded (dashed), achieved body-frame instantaneous velocity (actual), and average (mean) velocities as the crawling policy tracks forward, backward, and turning commands in one continuous run.}
    \label{fig:crawling_velocity_tracking}
    \vspace{-2pt}
\end{figure}

We evaluate velocity tracking in simulation by sampling piecewise-constant twist commands and holding each command for three gait cycles. Fig.~\ref{fig:crawling_velocity_tracking} compares the commanded twist with the instantaneous (pelvis) velocity and its mean over each command interval. Although the instantaneous velocities exhibit large phase-dependent oscillations due to the crawling gait, their cycle-averaged values follow the commanded trends across $v^x$, $v^y$, and $\omega^z$, as intended.
%
\subsection{Ablations and Method Comparison} \label{sec:ablation_comp}
We perform an ablation study and directly compare against existing methods to assess both our method's ability to produce trajectories advantageous for motion-tracking RL training and the performance of the resulting policies. All policies are trained across five random seeds using default mjlab motion-tracking settings \cite{mjlabtrackingsettings}, except that the termination gates and domain-randomization ranges are adjusted to match those used for hardware deployment. Policy evaluation is performed in a separate MuJoCo simulator featuring asynchronous control and sensing as well as randomized starting joint configurations. We report both RL training performance and sim-to-sim base pose tracking error\footnote{Positional error component of the pose is normalized by the robot's height and the orientation error component by $180^\circ$.} and joint tracking error. \rev{In Figs.~\ref{fig:ablation1} and~\ref{fig:ablation2}, $\sigma$ denotes the standard deviation of the PPO policy's exploration noise.}

\subsubsection{Dynamic Fidelity Ablation}\label{sec:dyn_ablation}

\begin{figure*}[t!]
    \centering
    \begin{minipage}[t]{0.48\textwidth}
        \centering
        \includegraphics[width=\linewidth]{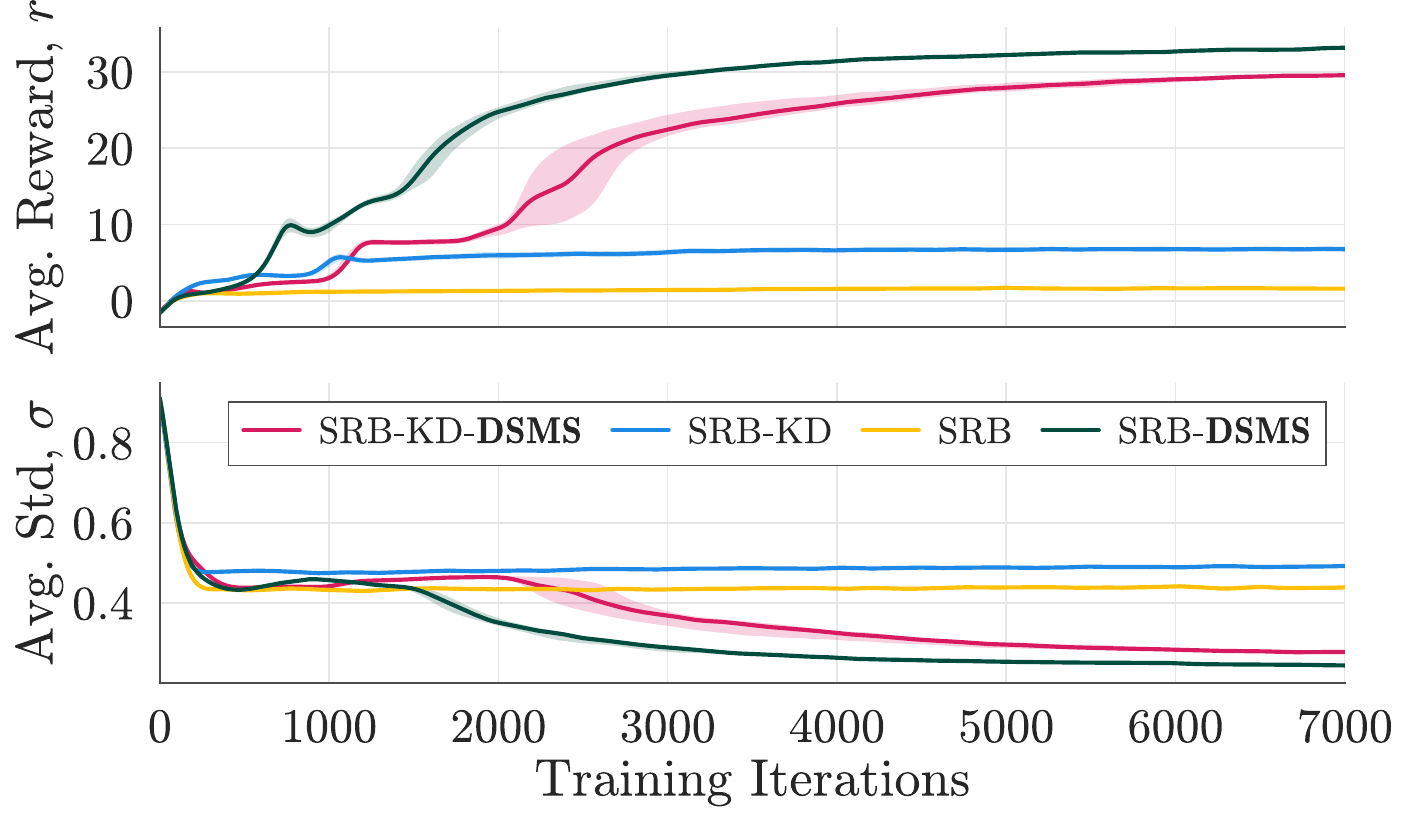}
        \caption{Training performance of motions generated from the ROM ablation, with variance (shade) across seeds.}
        \label{fig:ablation1}
        \vspace{-2pt}
        \footnotesize
        \setlength{\tabcolsep}{2pt}
        \setlength{\aboverulesep}{0pt}
        \setlength{\belowrulesep}{0pt}
        \renewcommand{\arraystretch}{1.15}
        \captionof{table}{ROM ablation sim-to-sim tracking accuracy.}
        \label{tab:Ablation_srb}
        \begin{tabular}{l|cccc}
            \toprule
            \textbf{Method} & \textbf{Landed} & \textbf{Success} & \textbf{Joint Err. (rad)}\textsuperscript{\dag} & \textbf{Pose Err.}\textsuperscript{\dag} \\\midrule
            SRB                                              & 0/75   & 0\%      & --               & --              \\
            SRB$\rightarrow$KD                             & 0/75   & 0\%      & --               & --              \\
            SRB$\rightarrow$\textbf{DSMS}               & \textbf{75/75}  & \textbf{100}\%     & \scalebox{0.92}{$\mathbf{0.110\pm0.006}$}   & \scalebox{0.92}{$\mathbf{0.155\pm0.041}$}     \\
            
            SRB$\rightarrow$KD$\rightarrow$\textbf{DSMS} & 72/75  & 96.0\%     & $0.144\pm0.016$    & $0.253\pm0.139$     \\
            \bottomrule
        \end{tabular}
    \end{minipage}
    \hfill
    \begin{minipage}[t]{0.48\textwidth}
        \centering
        \includegraphics[width=\linewidth]{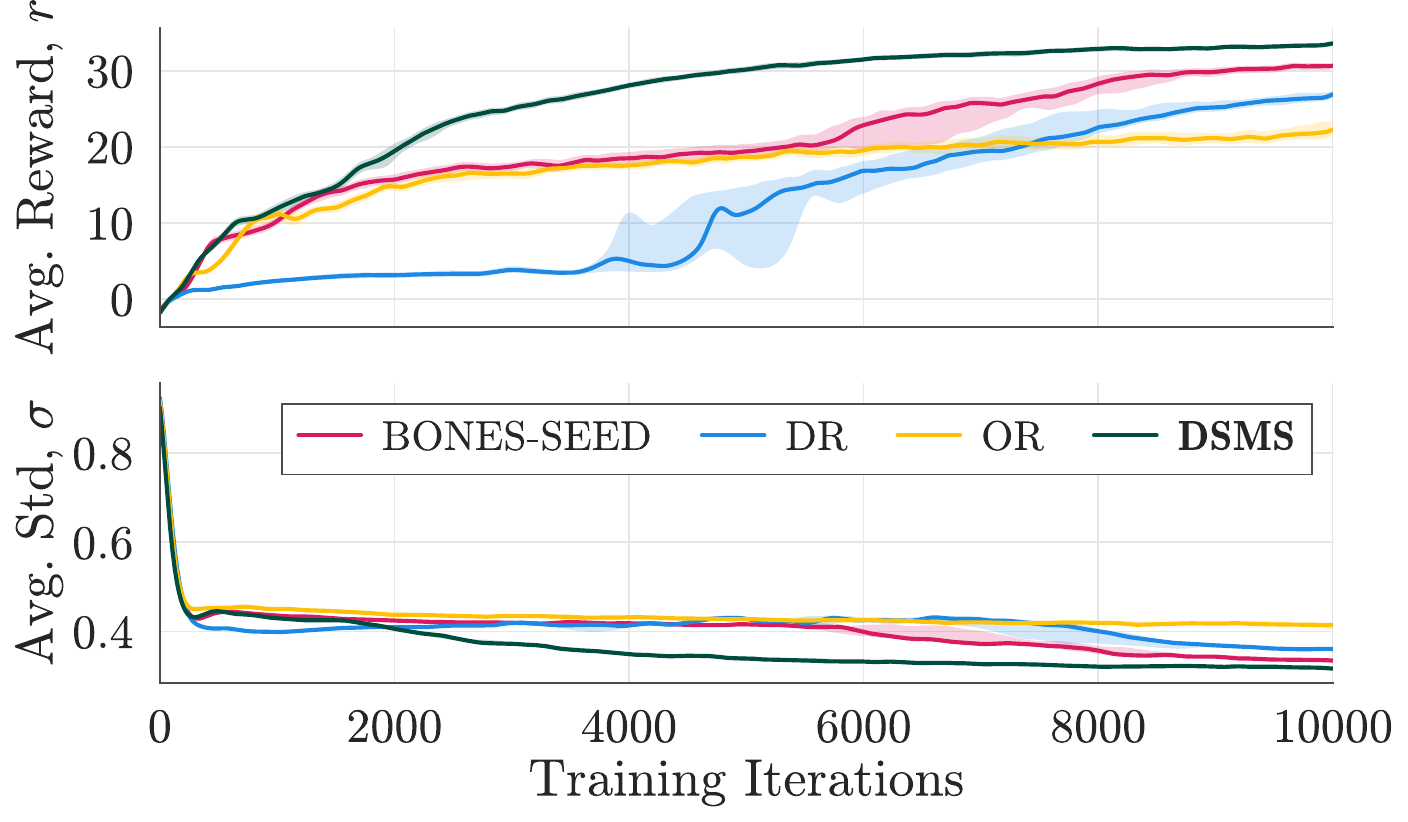}
        \caption{Training performance of super hero backflip motion across various retargeting methods.}
        \label{fig:ablation2}
        \vspace{-2pt}
        \footnotesize
        \setlength{\tabcolsep}{4pt}
        \setlength{\aboverulesep}{0pt}
        \setlength{\belowrulesep}{0pt}
        \renewcommand{\arraystretch}{1.15}
        \captionof{table}{Retargeting method tracking error comparison.}
        \label{tab:retarget_ablation}
        \begin{tabular}{l|cccc}
            \toprule
            \textbf{Method} & \textbf{Landed} & \textbf{Success} & \textbf{Joint Err. (rad)}\textsuperscript{\dag} & \textbf{Pose Err.}\textsuperscript{\dag} \\
            \midrule
            OR            & 7/75  & 9.3\%  & $\mathbf{0.157 \pm 0.001}$ & $\mathbf{0.095 \pm 0.014}$ \\
            BS            & 60/75 & 80.0\% & $0.195 \pm 0.003$ & $0.146 \pm 0.027$ \\
            DR            & \textbf{74/75} & \textbf{98.7\%} & $ 0.237 \pm 0.006$ & $0.160 \pm 0.077$ \\
            \textbf{DSMS} & \textbf{74/75} & \textbf{98.7\%} & $\mathbf{0.159 \pm 0.006}$ & $\mathbf{0.102 \pm 0.046}$ \\
            \bottomrule
        \end{tabular}
        \vspace{2pt}
        {\footnotesize $^{\dag}$~Errors only computed over successful landings.}
    \end{minipage}
    \vspace{-20pt}
\end{figure*}
This ablation studies DSMS efficacy when applied to trajectories generated using ROMs of increasing levels of dynamic fidelity. We begin with a backflip motion generated with an SRB ROM. Per-frame inverse kinematics (IK) then provides a fully kinematically feasible trajectory serving as a baseline for this ablation. A kino-dynamic (KD) optimization is performed on this trajectory to further refine the motion \cite{zhang2026kinodynamic}. Finally, our method can be applied at any stage to provide full dynamic feasibility, before all trajectories are passed to RL training for evaluation. The trajectory generated by SRB~$\rightarrow$~\textbf{DSMS} is shown in the backflip sequence of Fig.~\ref{fig:simulations}a.  
%
%

Fig.~\ref{fig:ablation1} and Table~\ref{tab:Ablation_srb} report the policy training performance for each trajectory and the tracking accuracy of the resulting policies. The two trajectories retargeted with DSMS converge fastest---indicated by lower PPO action noise $\sigma$---and achieve the highest reward. Sim-to-sim deployment supports that, under standard RL with minimal tuning, policies trained on dynamically-feasible motions outperform those trained on kinematic-only references.
%
\begin{table}[b]
	\centering
	\caption{Qualitative feature comparison across methods.}
	\label{tab:Ablation_capability}
	\footnotesize
	\setlength{\tabcolsep}{3pt}
	\setlength{\aboverulesep}{0pt}
	\setlength{\belowrulesep}{0pt}
	\renewcommand{\arraystretch}{1.15}
	\begin{tabular}{l|cccc}
		\toprule
		\textbf{Attribute} & \textbf{OmniRetarget} & \textbf{DynaRetarget} & \textbf{SPARK} & \textbf{Ours} \\
		\midrule
		Kinematic Feas.    & \checkmark            & \checkmark            & \checkmark     & \checkmark    \\
		WB Dyn. Feas.   & \ding{55}             & \checkmark            & \ding{55}      & \checkmark    \\
		Contact-Implicit   & \checkmark            & \checkmark            & \ding{55}      & \checkmark    \\
		Arbitrary Const.   & \ding{55}             & \ding{55}             & \checkmark     & \checkmark    \\
		Solver             & Seq. SOCP             & SBTO                  & Unspecified  & IPOPT         \\
		\bottomrule
	\end{tabular}
	\vspace{-5pt}
\end{table}
\subsubsection{Retargeting Method Comparison}
We compare against popular retargeting and optimization methods using a ``super hero backflip" (Fig.~\ref{fig:simulations}b) from BONES-SEED (BS) as a shared reference motion. This flip features contact across the arms, knees, and feet upon landing, as well as artifacts such as unrealistic knee clipping into the ground. The BS baseline is pre-retargeted to the G1, then dynamically retargeted with DSMS and also retargeted with OmniRetarget (OR) and DynaRetarget (DR) \cite{yang2025omniretarget, dhedin2026dynaretarget}. Each resulting trajectory is then used to train motion-tracking policies for evaluation, as discussed previously.

As shown by the training curves (Fig.~\ref{fig:ablation2}), the trajectory optimized with DSMS again converges fastest, reaching an average reward of 20 in ${\sim}2000$ iterations versus the ${\sim}5000$ required by its closest competitors---a 40-minute wall-clock difference on an Nvidia RTX 4090. Given our method's ${\sim}12$-minute optimization time, this advantage is worthwhile even before accounting for the runtime of competing methods. The combined sim-to-sim tracking metrics (Table~\ref{tab:retarget_ablation}) tell the same story: DSMS attains the highest backflip success rate, near-lowest tracking error, making it the only method to consistently perform the hero backflip with accurate tracking.



\subsection{Qualitative Capability Comparison}\label{sec:features}
To augment our discussion of retargeting and trajectory optimization approaches and situate DSMS among similar approaches, we present Table~\ref{tab:Ablation_capability}.

Here, whole-body dynamic feasibility refers to a trajectory satisfying \eqref{eq:whole_body_dynamics}, while arbitrary constraints refers to a method's ability to enforce constraints of the form of \eqref{eq:ocp_inequality} and \eqref{eq:ocp_equality}. Although all four methods provide kinematic feasibility, they differ substantially in the physical and constraint-handling ability. OmniRetarget achieves excellent kinematic retargeting and often produces visually plausible trajectories, but it sacrifices dynamic feasibility for kinematic accuracy. SPARK shares this limitation, and although it can incorporate arbitrary constraints, it is not contact-implicit: it relies on holonomic constraints that require accurate contact detection for the solver to function. DynaRetarget instead samples roll-outs to incorporate dynamics and handle contact implicitly\rev{; however, it is} not able to accommodate arbitrary constraints.

DSMS combines whole-body dynamic feasibility with a contact-implicit formulation, enabling contact-rich motions such as rolling and crawling without extensive tuning. It also supports arbitrary equality and inequality constraints. Our ablations further show that better-modeled trajectories yield better downstream tracking controllers, underscoring the importance of this approach.

%% file: Sections/Conclusion.tex
\section{Conclusion}


We presented a contact-implicit multi-shooting framework for transforming kinematically feasible references into dynamically feasible whole-body humanoid motions. By using a differentiable simulator as the transition model, the framework accounts for contact, friction, impacts, self-collision, and actuation limits without requiring a prescribed contact sequence or manual constraint setup. The resulting reference trajectories improve downstream RL convergence and tracking performance while enabling contact-rich and dynamic behaviors such as \rev{walking, jogging,} crawling, jumping, backflip, and side-rolling. 

Future work will investigate alternative differentiable simulators and sampling-based optimization methods, automatic discovery of contact-rich behaviors, \rev{robot--object interaction,} and extension to additional robot morphologies.